# Multi-Modal Emotion Recognition by Text, Speech and Video Using Pretrained Transformers

Bagher Babaali, Minoo Shayaninasab

**Abstract**

Due to the complex nature of human emotions and the diversity of emotion representation methods in humans, emotion recognition is a challenging field. In this research, three input modalities, namely text, audio (speech), and video, are employed to generate multimodal feature vectors. For generating features for each of these modalities, pre-trained Transformer models with fine-tuning are utilized. In each modality, a Transformer model is used with transfer learning to extract feature and emotional structure. These features are then fused together, and emotion recognition is performed using a classifier. To select an appropriate fusion method and classifier, various feature-level and decision-level fusion techniques have been experimented with, and ultimately, the best model, which combines feature-level fusion by concatenating feature vectors and classification using a Support Vector Machine on the IEMOCAP multimodal dataset, achieves an accuracy of 75.42%.

## 1. Introduction

Most research in the field of emotion recognition do not focus on integrating different models into a system for analyzing human emotional behavior. Instead, various channels of emotional information are examined independently. Nevertheless, an ideal system for analyzing and detecting human emotional information should be multimodal and operate similar to the human sensory system.

In the field of unimodal emotion recognition, extensive research has been conducted using various modalities. Facial expressions, vocal features, body movements, and body positions, as well as physiological signals, have been used as outputs in these systems. Nowadays, multimodal emotion recognition is advancing rapidly. Most studies focus on combining information from facial expressions and speech. Various combinations of modalities have been proposed in the literature, including the combination of facial expressions with speech, audio-visual signals, physiological signals, and EEG responses with textual content [1].

Multimodal learning is a type of learning that offers greater efficiency compared to unimodal learning. However, the use of multimodal systems also comes with its challenges. For instance, selecting the modalities for fusion, dealing with missing data, lack of sufficient multimodal datasets, managing synchronization and integration of different modalities are some of the most important challenges in multimodal systems. Among these challenges, finding the best combination of modalities and discovering the optimal mechanism for fusing modalities have been the focus of recent research. Currently, multimodal learning techniques are used as a complementary method to enhance the accuracy of emotion recognition, particularly deep learning in multimodal emotion recognition. Deep learning techniques encompass deep belief networks, deep convolutional neural networks, long short-term memory, and combinations thereof. However, more recent studies often focus on using self-supervised pre-trained models, multi-task learning, attention mechanisms, autoencoders, and so on.

## 2. Related work

A wide variety of machine learning techniques have been employed in emotion recognition methods. Especially in multimodal scenarios, numerous audio, visual, or physiological features are often used, which can make the training process challenging. Therefore, it is essential to find a way to reduce the number of features used. One of the solutions is the use of neural networks, as these networks typically allow us to discover relevant features related to the output by observing their weights.

Recently, the combination of audio and visual modalities has achieved significant success in emotion recognition. Some studies have leveraged useful features that deep neural networks can extract.

In the research [2], experiments have been conducted on the IEMOCAP multimodal dataset to compare the system's performance under conditions where it encounters new subjects in the test set versus conditions where the people in the test and train datasets overlap. In this work, a convolutional neural network is used for extracting textual features, which are n-gram features obtained using three distinct convolutional filters of sizes 3, 4, and 5. For extracting audio and visual features, openSMILE and a 3D convolutional network are employed, respectively. The research was conducted on four classes of data from the IEMOCAP dataset. The workflow involved extracting features and feeding them to a support vector machine for classification. An innovation in this approach is the use of a bidirectional long short-term memory model for extracting context information.

In [3], to build a multimodal system on the IEMOCAP dataset, various architectural proposals were initially presented and tested for each modality, and ultimately, the architectures that yielded the best results for each modality were combined in the multimodal system. In this research, instead of processing video with a 3D convolutional network, a model was used to extract motions first, and then these motions were processed in a 2D convolutional network, which made the training process faster and lighter while maintaining nearly constant accuracy compared to state-of-the-art emotion recognition systems.

In previous works, convolutional neural networks and long short-term memory networks have been used as fusion mechanisms. Recently, the use of transformers for multimodal emotion recognition has garnered significant attention. The transformer is a network architecture that relies entirely on the attention mechanism without any recurrent structure. Recent research has focused on using attention mechanisms to combine features from different modalities for multimodal emotion recognition.

Ho and his colleagues [4] presented a multimodal approach based on multi-head attention and recurrent networks to combine audio and textual modalities for emotion prediction. Furthermore, the paper suggests that the attention mechanism fusion method improves the performance of emotion recognition in the multimodal setting compared to unimodal approaches.

various experiments are conducted on emotion recognition models. In [5], modifications to the multi-head attention mechanism formula have been made in a way that allows the creation of two different systems in the fusion stage. This study employed a transformer for combining audio and visual modalities at the multi-head attention level. They compared decision-level fusion with feature-level fusion using the attention mechanism. The experiments showed that the proposed approach through feature-level fusion achieved superior improvements.

In [6], three different modalities, text, audio, and images, with features extracted from pre-trained self-supervised learning transformer models were combined. This study designed a multimodal fusion mechanism that employs six transformer blocks on single-modal transformer models to integrate the information from different modalities by applying cross-attention across the interacting modalities. In essence, in the self-attention formula of these blocks, the queries and keys come from each pair of modalities that are supposed to be fused. This fusion process involves a complex training process but ultimately achieved state-of-the-art results.

While the effectiveness of combining two different modalities using the attention mechanism has been extensively studied, a challenge arises when the need for combining three or more modalities arises due to the multi-head attention mechanism's structure. Therefore, most previous studies on multimodal emotion recognition based on the attention mechanism have only presented and tested network architectures for two modalities. However, in practice, fusing different high-dimensional features may lead to sparse data and reduced performance.

## 3. Experimental setup

### 3.1. Data preparation

This research uses IEMOCAP [7] dataset to train and evaluate an emotion recognition system. First, a good understanding of the IEMOCAP dataset is required. Each modality has 10,039 data points with 11 imbalanced labels. The data points that falls into four emotional categories: neutral, anger, sadness, and happiness are chosen. The data used in this study consists of 5,532 data rows. There is also an emotion label, 'excited', with a low frequency, but due to the similarity of emotional features to happiness, data from its subset has been selected in the happiness category. The frequency distribution of each label in each session of the selected data is shown in Figure 1.

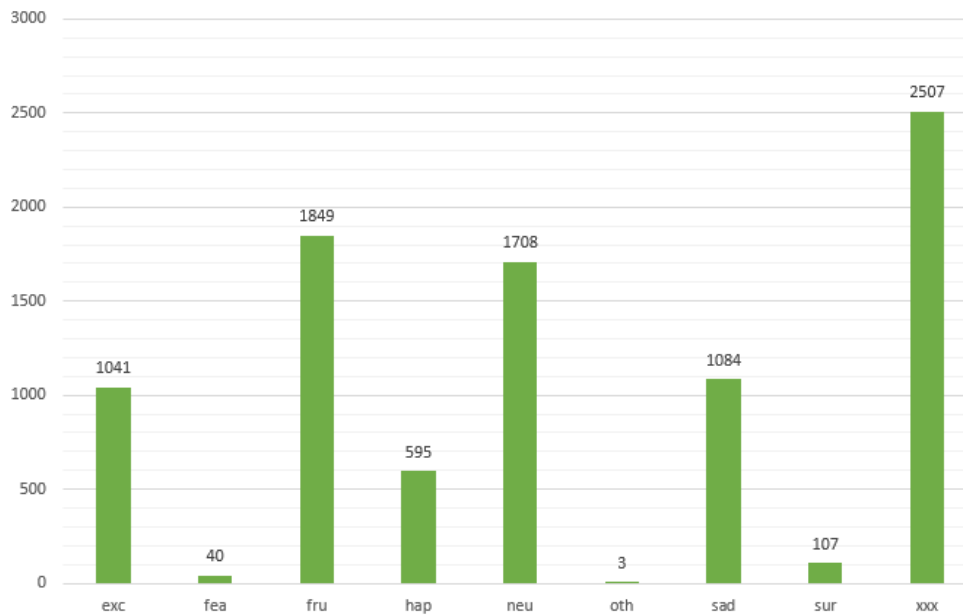

Figure 1: Distribution of Selected Labels in Five Sessions

An optimal data loading method is needed. For text, each sentence or turn from the file that contains that conversation is read and separated. Handling audio data is simpler since the segmentation in the speech for each sentence in the original dataset is available, and the only necessary task is matching them with labels. To achieve this, the turn number of the sentence is considered as the primary key. Raw videos need to be segmented to extract video clips for each sentence. The timing information for each turn is extracted from the text files, and the videos are cut accordingly. The output videos are converted to MP4 while preserving the maximum possible quality. Since sound is not required for training the visual (image) processing networks in this research, the audio is removed from these videos to reduce storage space and computation load.

Preparing data and dataloaders was crucial as the existence of an efficient data pipeline makes the training process significantly faster and easier. It is essential for the data pipeline method to be versatile, compatible with various models and libraries, to encounter fewer issues during training.

### 3.2. Model selection

In the next section of the work, three text models and three speech models need to be fine-tuned on the data to select the model that performs the best for the fusion step. To accomplish this, the data is split into three sessions: the first three sessions are allocated for training, the fourth session is designated for validation, and the fifth session is assigned for testing. These models are fine-tuned by adding a fully connected layer and a softmax layer on top of each of them. These additional layers are employed for single-modal classification.

For the selection of a text model, three models, namely BERT [8], RoBERTa [9], and DistilBERT [10], are loaded from Hugging Face. The operations in this section are executed using a combination of PyTorch and Transformers functions. Tokenization and automatic length normalization of the outputs are taken care of by the model class functions. The accuracy metric is defined, along with a function for calculating the cross-entropy loss. Each model is fine-tuned with a learning rate of 2e-5. The batch size for training and validation is set to 32. To enhance model convergence and prevent overfitting, a weight decay algorithm with a fixed parameter of 0.01 is employed. Training is scheduled for 20 epochs, and the best model based on the validation accuracy is chosen and saved. The accuracy of all three fine-tuned text models on the test dataset is calculated.

As for the speech model, three models, namely wav2vec2.0 [11], HuBERT [12], and Audio Spectrogram Transformer [13], are fine-tuned on the speech data. The data division follows a similar pattern to what was used for the text models. To preprocess the outputs, a 16 kHz sample rate is maintained. The maximum audio input length is set to 10 seconds, and longer outputs are sliced. Feature extraction is carried out using auxiliary functions from Transformers and PyTorch. In the training section, a learning rate of 3e-5 is defined. The training for HuBERT and wav2vec 2.0 continues for 15 epochs, while the Audio Spectrogram Transformer trains for 30 epochs. The batch sizes for training and validation for these models are set to eight and four, respectively. A weight decay algorithm with a fixed parameter of 0.01 is applied.

To ensure the performance of the video model, the videoMAE [14] model is fine-tuned on the prepared video data. This fine-tuning is performed over 10 training epochs with a batch size of eight, a learning rate of 5e-5, and the same weight decay parameter of 0.01 as mentioned for the previous two sections. In the video data preprocessing, randomly mirrored videos with a 50% chance are used to increase data diversity. Videos are preprocessed and samples entering the network are sampled at 30 frames per second. The pixel colors are normalized. Some parts of this are implemented while most of it is done using the available methods in PyTorchVideo. The training steps, data pipelines, and evaluation procedures are conducted similarly to the text and speech models, with Transformers and PyTorch.

### 3.3. Fusion

By comparing the initial accuracy of text and speech models on the test sets, BERT and wav2vec2.0 are selected for the fusion stage. Since the final model results are reported through 5 fold cross-validation, before fusion, five models are prepared for speech, five models for text, and five models for video, using the pretrained models BERT, wav2vec2.0, and videoMAE, with the same training parameters mentioned previously. In these 15 models, each time, one session is entirely considered for the test set. The data passes through these transformers and the representations are saved to be fused.

In this research, the output representations generated by transformer models in text, audio, and video are fused in two main approaches: early fusion and late fusion. In both approaches, the extracted vectors from each of the selected models are either summed, multiplied or concatenated to create a representation of the multi-modal data.

In late fusion, the classification results from the last layer of the single-modal models are fused. With early fusion, the decision layers of each single-modal model are completely removed, and the feature representations are extracted from layer preceding the final layer of the transformers. Early fusion is also called feature-level fusion because it behaves like single-modal models, extracting features from a layer of the deep network and incorporating the generated features as part of the data representation of the main model. In the typical transformer architecture, the CLS token at the last layer is expected to encode all the information needed for classification. This representation from the standard transformer architecture has a dimension of 768.

### 3.4. Classification

The result of the classification is the emotion recognition task. The multi-modal representations described above are fed into the classifier. Three classification methods, SVM, XGBoost and a 2-layer neural network are applied. The reason for using the average of three classifiers to evaluate the fusion method is to assess the performance independently of the classifier and its effectiveness.

The final model obtained from this research is a model that fuses BERT, wav2vec2.0, and videoMAE through early fusion by concatenating the representations of the three modalities and applies a Support Vector Machine (SVM) classifier.

## 4. Results and discussion

### 4.1. Training and accuracy

Results from initial trainings to choose between BERT, roBERTa and DistilBERT are reported in table 1. The only metric used to choose is the accuracy.

|  | Loss | Accuracy | Runtime |
|---|---|---|---|
| BERT | 1.19 | 0.68 | 4.28 |
| DistilBERT | 0.91 | 0.65 | 2.22 |
| roBERTa | 1.48 | 0.65 | 4.32 |

*Table 1: comparison between initial text models*

For the choosing between HuBERT, wav2vec2.0 and AST, accuracies from table 2 were measured. Also an initial training on videoMAE

was performed and the results are reported in table 3.

|  | Loss | Accuracy | Runtime |
|---|---|---|---|
| Wav2vec2.0 | 0.95 | 0.68 | 90.60 |
| HuBERT | 1.07 | 0.625 | 73.84 |
| AST | 3.07 | 0.61 | - |

Table 2: comparison between initial speech models

|  | Loss | Accuracy | Runtime |
|---|---|---|---|
| VideoMAE | 1.32 | 0.38 | 629.72 |

Table 3: initial video model

In order to report the accuracy of the selected single-modal models, fine-tuning is performed five times for each model. These results are reported in Table 4.

|  | Loss | Accuracy | Runtime |
|---|---|---|---|
| BERT | 1.27 | 67.50 | 3.75 |
| Wav2vec2.0 | 1.41 | 67.55 | 81.46 |
| VideoMAE | 1.35 | 33.22 | 562.62 |

Table 4: uni-modal performances

The multi-modal classification results after the fusion in performed is reported in table 5 and 6 which correspond to the early and late fusion. Among the three classifiers in table 5, Support Vector Machine has performed more accurately with a negligible difference from the two-layer neural network. In most previous studies, there is a significant difference between fusion methods. In this study, early fusion performed better, although the difference is very small. In the early fusion case, the sum of feature vectors worked more accurately compared to the other method.

| SVM | NN | XGBoost | Average |
|---|---|---|---|
| Concatenation of representation vectors | | | |
| 75.451 | 75.31 | 73.28 | **74.68** |
| Summation of representation vectors | | | |
| 75.42 | 75.16 | 73.95 | **74.84** |
| Multiplication of representation vectors | | | |
|  |  |  |  |

Table 5: results for early fusion

| SVM | NN | XGBoost | Average |
|---|---|---|---|
| Concatenation of representation vectors | | | |
| 73.97 | 74.36 | 74.45 | **74.26** |
| Summation of representation vectors | | | |
| 72.10 | 72.56 | 72.75 | **72.47** |
| Multiplication of representation vectors | | | |
|  |  |  |  |

Table 6: results for late fusion

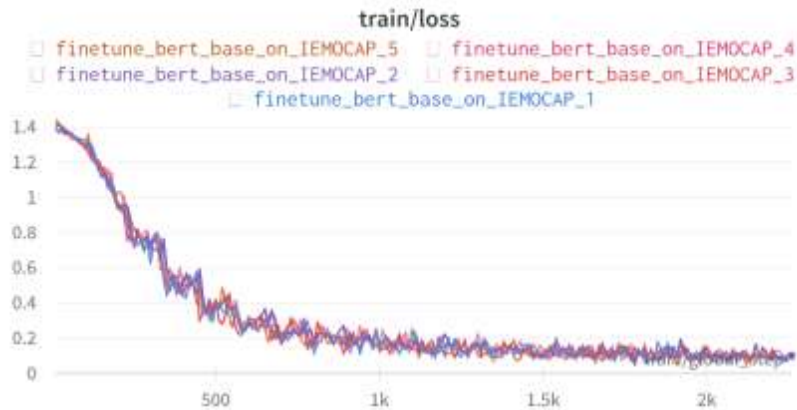

Figure 2: loss function by train steps for finetuning bert on IEMOCAP text data

The convergence plots of the training of the models prepared at this stage are shown in Figures 2, 3 and 4. The charts correspond respectively to the text, speech, and video modalities. For text and speech, the model errors on training data are very close to each other, and this holds true for the evaluation error as well. However, the text model exhibits slightly more consistency. Regarding videos, the situation is different. There is an issue with incomplete model training, leading to a significant difference in performance. As we move closer to the end regions, relative convergence occurs, and the models become more trained.

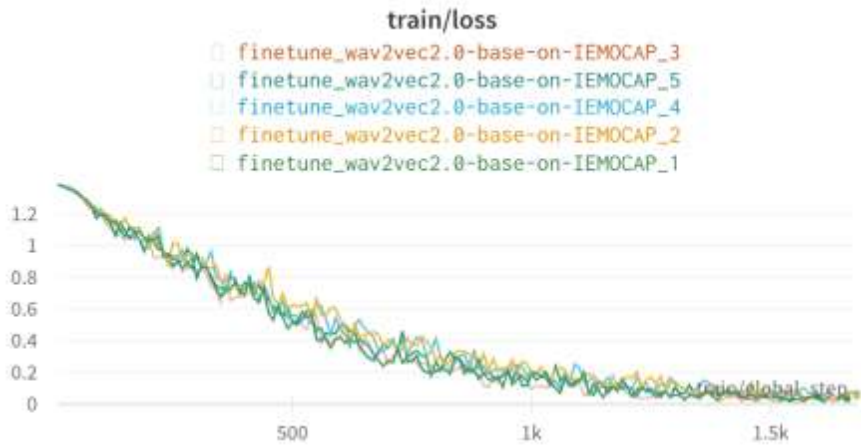

*Figure 3: loss function by train steps for finetuning wav2vec2.0 on IEMOCAP speech data*

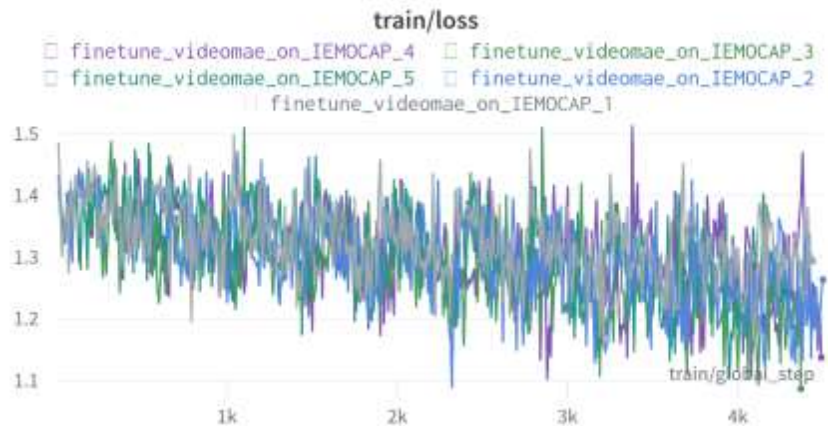

*Figure 4: loss function by train steps for finetuning videoMAE on IEMOCAP video data*

In Figure 5, 6 and 7 the classification accuracy of the five models created for 5-fold cross-validation on each of the modalities is displayed. The plots correspond to the text, speech, and video modalities respectively. The difference in the curves is indicative of the same previous results. Note the difference in the numerical range shown on the charts for different modalities. The video model's accuracy differs significantly from the other two modalities.

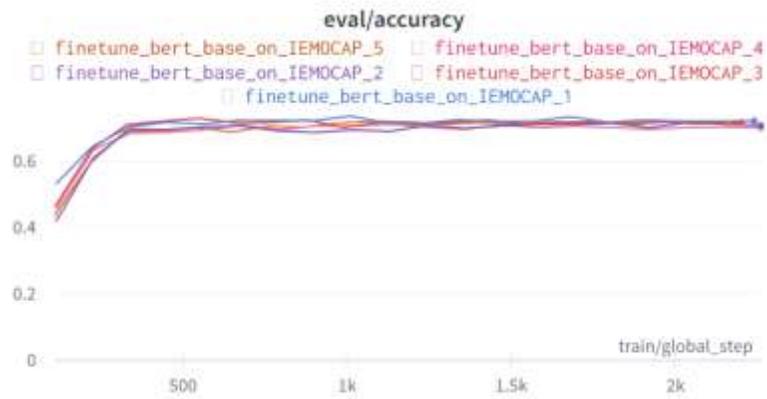

*Figure 5: text model accuracy*

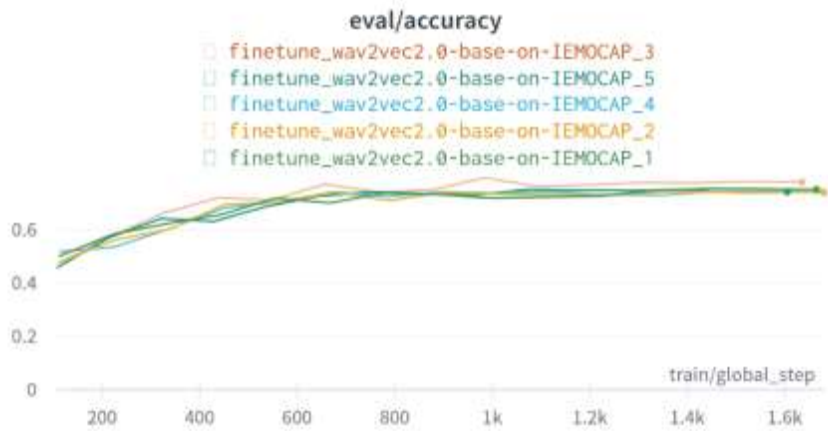

*Figure 6: speech model accuracy*

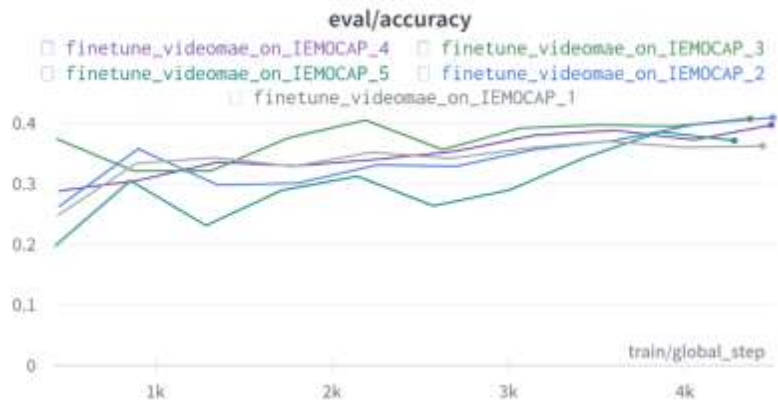

*Figure 7: video model accuracy*

## 4.2. Data visualization

Figures 8 to 10 depict the last layer representation of the CLS token of each of the unimodal models which are plotted in two-dimensional space. A PCA transformation is used since each unimodal representation vector had a dimension of 768. These representations are concated to create the multi-modal representations that are ploted in figure 11.

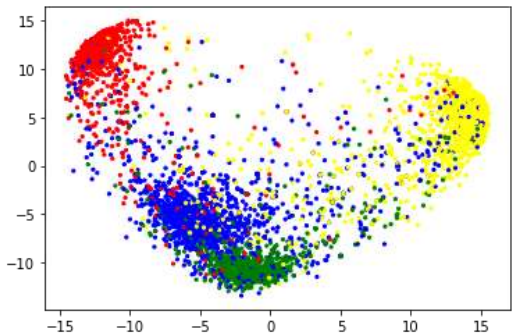

*Figure 8: 2-D plot of text data representations*

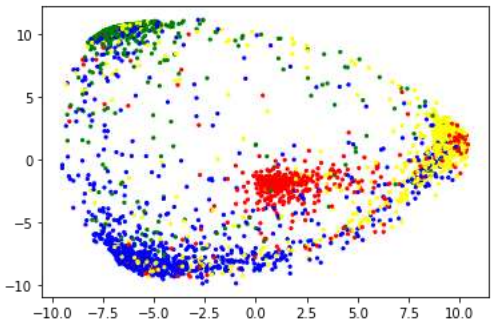

*Figure 9:2-D plot of speechdata representations*

As expected from the numerical data in the previous section, the models for text and speech have significantly higher capabilities for class separation compared to the video model. In the plots presented in figure 8-11, the colors red, yellow, blue, and green correspond to the classes anger, happiness, neutral, and sadness, respectively. In the feature space for text and speech, the four classes represented by four different colors, although intertwined, are somewhat separable within specific hypothetical regions, though not entirely. However, in the video feature space, there is no such clear separation.

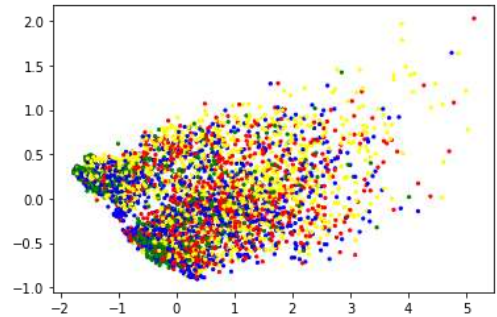

*Figure 10: 2-D plot of video data representations*

While it's possible to observe the density of green points in the left section of the plot, for the text model, most of the overlap is between the green and blue classes, indicating that in emotion recognition from text data, distinguishing between the sadness and neutral classes are the most significant challenge. For the speech model, there is more separation between the classes, but the blue class still exhibits a significant amount of spread in the feature space. Detecting the neutral or no-emotion class is likely a primary factor contributing to the error in the speech model. Both anger and happiness classes, particularly anger in the text model, seem to have the best separation from the other classes.

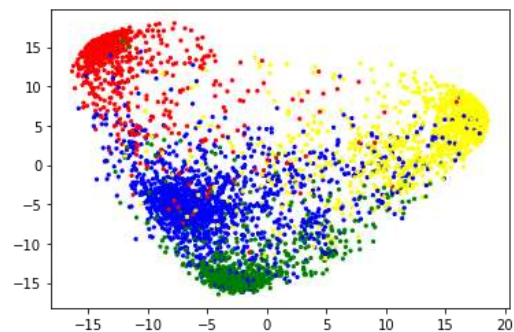

*Figure 11: 2-D plot of multi-modal data representations*

From the numerical data in the previous section, one might expect that combining these two models could compensate for each other's weaknesses. However, the fusion of these two models with the video model, as seen in Figure 11, is an action aimed at classifying video data. The last graph clearly shows that the fusion of the modalities is an effective step. This visual representation shows the improved discriminative power of multimodal models compared to unimodal models.

### 4.3. More analysis

In this section, the obtained model is subjected to new metrics to better understand its strengths and weaknesses. Table 7 displays precision, recall, and F1-score. To further analyze the results in this section, we will consider observations of data visualized in a two-dimensional space. A superficial glance at the third column of Table 7 reveals that while the F1-score across all data is acceptable, these scores vary among the different classes. The weakest F1-score is obtained for the neutral class. Both recall and precision for this class are lower compared to other classes.

|         | Precision | Recall | F1-score |
|---------|-----------|--------|----------|
| Angry   | 79.52     | 85.66  | **82.33** |
| Happy   | 77.75     | 75.72  | **76.46** |
| neutral | 70.19     | 69.41  | **69.27** |
| sad     | 78.41     | 74.35  | **75.69** |
|         | 76.10     | 75.42  | **75.35** |

*Table 7: precision, recall and F1 for multi-modal emotion recognition*

Relooking the 2-D data visualization, the blue class corresponding to the neutral category has spread significantly among different classes. Data points that move away from the center of the neutral class area are mistakenly classified into other classes, reducing the recall for the neutral class. Furthermore, the precision of other classes is also reduced. Essentially, the F1-score, which is a balance between precision and recall, indicates how confident the model's predictions are for a specific category. The best F1-score belongs to the anger class, which has both higher precision and recall. A high recall for the anger class suggests that this class is more reliably identified by the model.

To further investigate the strengths and weaknesses identified in the F1-score, it is better to examine a lower-level evaluation metric. In the confusion matrix, the number of data points that are correctly and incorrectly classified are presented. To present the confusion matrix in a more suitable way, a heat map is used, as shown in Figure 12.

Among the rows, the most prominent color area belongs to the neutral class, as expected from the previous analysis. However, the intention here is to understand in more detail which classes are causing more errors due to overlap. The neutral class is often mistakenly classified as the happy class, and vice versa. The sadness class follows with a higher number of misclassifications with the neutral class. Considering the preliminary analysis performed before the modeling stage regarding the distribution of data classes, there is a greater need to control errors related to the neutral class because this class is the most frequent among the four selected classes in the problem. This class does not represent a specific emotional label; instead, it reflects the absence of a distinct emotion. In other words, mistakes in recognizing this class result in a drop in the system's performance in recognizing emotions, which are expected to have different characteristics learned.

Continuing the examination of the heat map, the anger class exhibits the best separation from the other classes, and even fewer mistakes are made between it and the neutral class. The model performs excellently in recognizing anger. The sadness class has a great separation from both the happy and anger classes, and the expectation is that the errors concerning neutrality in this class are fewer compared to the happy class. However, there is still room for further investigation for improvement.

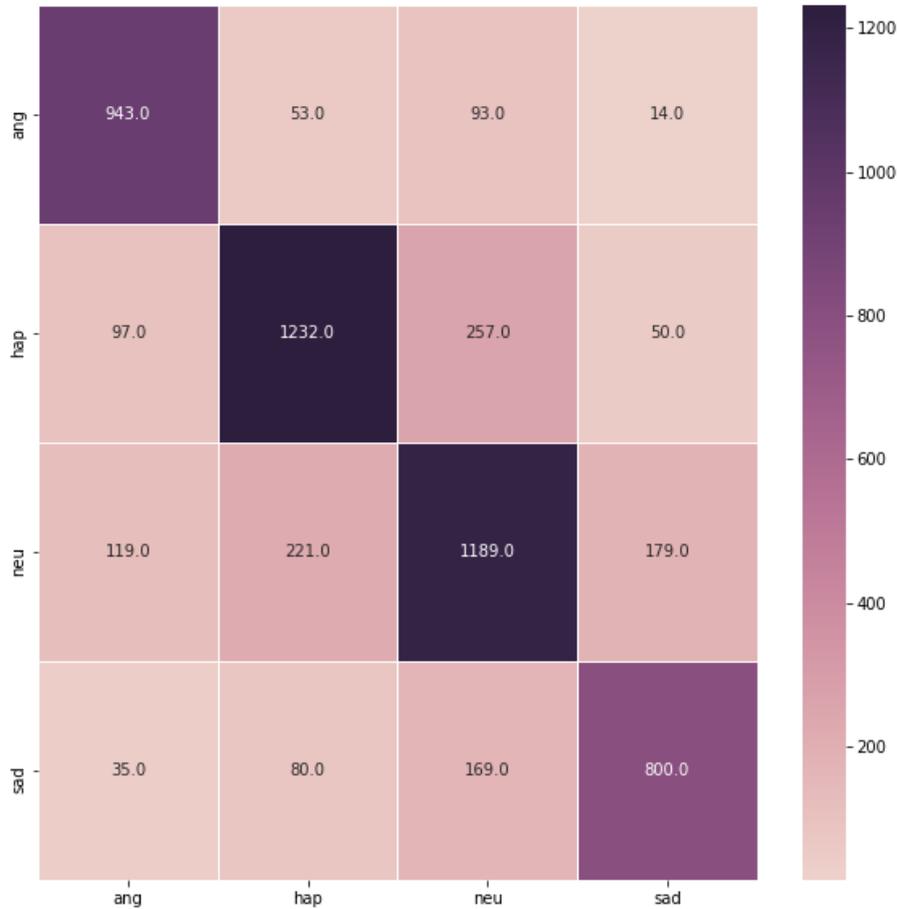

*Figure 12: heatmap of the confussion matrix*

The comparison of the emotion recognition research in [2] with the final model developed in our research highlights a couple of key points. Firstly, it's evident that the accuracy achieved here (75.42%) surpasses the model that was developed in [2] research (71.59%), despite not explicitly modeling background information for each modality. The proximity in performance between the model (75.42%) and the model with background information (76.1%) in [2] suggests that the transformer-based approach is capable of modeling the background information effectively due to its high feature extraction complexity. The second point is that while the enhanced single-modal models with domain knowledge yielded better results than our model in text and video, the overall performance is quite close. In the single-modal scenarios for text and video, using stronger models would likely further improve the results for the multi-modal model as well. The comparison of the audio model with the transformer architecture used for speech indicates that the transformer architecture can indeed model data effectively. Therefore, there is potential for further development in text and video by employing such architectures.

While the weakest and strongest performance of f1-score in our models align with the expectations we've had, the transformer-based fusion approach proposed in [6] which was previously discussed in section 2 has significantly improved the results due to the innovative fusion method.

5. **Conclusion**

This research has demonstrated that utilizing transfer learning for emotion recognition, especially when dealing with limited data, is a highly suitable approach. The accuracy of the final model indicates that transformers, are capable of addressing the challenges in modeling contextual information for emotion recognition. This is significant since, one major factor contributing to the errors in emotion recognition systems is the lack of sufficient data. Another factor that contributes to the superiority of transformers in emotion recognition is the automation of data preprocessing and preparation methods. When comparing the modeling process in this research with feature extraction methods introduced in the literature, it is evident that the automated procedures offer a more robust and efficient workflow.

Transformers introduced in for speech and text had very similar results, unlike other models. This suggests that, although an optimal transformer model can be found, the difference in performance is not substantial between different transformer models. The comparison of fusion methods shows the superiority of feature-level fusion. Another important consideration is the optimal size of the multimodal representation vector, which should not be too large or too small. An excessively large size does not yield desirable results while a very small size reduces the richness of the representation. This helps to design the architectures of models correctly.

The third factor explored in this research was to examine the hypothesis that if single-modal models have significantly higher power and complexity than traditional models, whether the fusion of information in a multimodal system can still increase accuracy. Although the difference between the performance of multimodal and single-modal systems in this context is less pronounced than in older research, it is still significant. The assumption that one modality cannot convey all sensory data remains valid.

Among the various models for different modalities, video models exhibited more significant weaknesses in classification. Training video models, particularly transformers, demands substantial computational resources and time. Memory limitations can lead to training with smaller batches, which can slow down model convergence. Also by further examining other similar emotion recognition tasks and the current applications of video transformers, it becomes evident that video transformers have not yet achieved the necessary power relative to their text and audio counterparts to surpass traditional methods. However, substantial advancements are expected in the near future.

Finally, it should be noted that emotion recognition, especially in the multimodal context, faces challenges due to a lack of suitable datasets. Current developed systems, such as the one discussed here, although promising, do not yet possess the necessary generalizability for practical applications. Emotions are highly diverse across languages, cultures, and human psychological states. It is natural that with just a handful of sensory datasets, we cannot expect results that can be generalized to the entire scope of this issue. Therefore, models developed in this context are initial steps for tackling a larger and more complex problem that requires diverse and abundant data.

This research is an attempt to benchmark the power of transformers in multi-modal emotion recognition and a system was built successfully with the accuracy of 75.42% on the IEMOCAP. The experiments here aimed to compare two-tier feature fusion and decision fusion methods. These results can be useful for other tasks requiring data fusion. Many challenges and limitations were also addressed to be considered in future research in the field of emotion recognition.

## 6. References

[1] Caridakis, George, et al. "Multimodal emotion recognition from expressive faces, body gestures and speech." Artificial Intelligence and Innovations 2007: from Theory to Applications: Proceedings of the 4th IFIP International Conference on Artificial Intelligence Applications and Innovations (AIAI 2007) 4. Springer US, 2007.

[2] Poria, Soujanya, et al. "Multimodal sentiment analysis: Addressing key issues and setting up the baselines." IEEE Intelligent Systems 33.6 (2018): 17-25.

[3] Tripathi, Samarth, Sarthak Tripathi, and Homayoon Beigi. "Multi-modal emotion recognition on iemocap dataset using deep learning." arXiv preprint arXiv:1804.05788 (2018).

[4] Ho, Ngoc-Huynh, et al. "Multimodal approach of speech emotion recognition using multi-level multi-head fusion attention-based recurrent neural network." IEEE Access 8 (2020): 61672-61686.

[5] Huang, Jian, et al. "Multimodal transformer fusion for continuous emotion recognition." ICASSP 2020-2020 IEEE International Conference on Acoustics, Speech and Signal Processing (ICASSP). IEEE, 2020.

[6] Siriwardhana, Shamane, et al. "Multimodal emotion recognition with transformer-based self supervised feature fusion." IEEE Access 8 (2020): 176274-176285.

[7] Busso, Carlos, et al. "IEMOCAP: Interactive emotional dyadic motion capture database." Language resources and evaluation 42 (2008): 335-359.

[8] Devlin, Jacob, et al. "Bert: Pre-training of deep bidirectional transformers for language understanding." arXiv preprint arXiv:1810.04805 (2018).

[9] Conneau, Alexis, et al. "Unsupervised cross-lingual representation learning at scale." arXiv preprint arXiv:1911.02116 (2019).

[10] Sanh, Victor, et al. "DistilBERT, a distilled version of BERT: smaller, faster, cheaper and lighter." arXiv preprint arXiv:1910.01108 (2019).

[11] Baevski, Alexei, et al. "wav2vec 2.0: A framework for self-supervised learning of speech representations." Advances in neural information processing systems 33 (2020): 12449-12460.

[12] Hsu, Wei-Ning, et al. "Hubert: Self-supervised speech representation learning by masked prediction of hidden units." IEEE/ACM Transactions on Audio, Speech, and Language Processing 29 (2021): 3451-3460.

[13] Gong, Yuan, Yu-An Chung, and James Glass. "Ast: Audio spectrogram transformer." arXiv preprint arXiv:2104.01778 (2021).

[14] Tong, Zhan, et al. "Videomae: Masked autoencoders are data-efficient learners for self-supervised video pre-training." Advances in neural information processing systems 35 (2022): 10078-10093